\documentclass{article}
\usepackage{spconf,amsmath,graphicx}

\usepackage{microtype}
\usepackage{multirow}
\usepackage{makecell}
\usepackage{soul} 
\usepackage{color, xcolor} 
\usepackage{inconsolata}
\usepackage{times}
\usepackage{latexsym}
\usepackage{microtype}
\usepackage{amsmath}
\usepackage{booktabs}
\usepackage{multicol}
\usepackage{multirow}
\usepackage{graphicx}  
\usepackage{float}  
\usepackage{subfigure}  
\usepackage{algorithm}
\usepackage{algorithmic}

\title{Empathetic Response Generation via Emotion Cause Transition Graph}
\name{
    \begin{tabular}{c}
        Yushan Qian$^{\dagger \ddagger \heartsuit}$, Bo Wang$^{\dagger*}$, Ting-En Lin$^{\ddagger}$, Yinhe Zheng$^{\ddagger}$ \\ 
        Ying Zhu$^{\dagger}$, Dongming Zhao$^{\dagger}$, Yuexian Hou$^{\dagger}$, Yuchuan Wu$^{\ddagger}$, Yongbin Li$^{\ddagger*}$
    \end{tabular}
    \thanks{$^{*}$ Corresponding authors.}
    \thanks{$\heartsuit$ Work done while interning at Alibaba.}
}
\address{
    $^{\dagger}$State Key Laboratory of Communication Content Cognition, People’s Daily Online, Beijing, China \\ 
    $^{\ddagger}$Alibaba Group, Beijing, China \\
    \texttt{shuide.lyb@alibaba-inc.com}
}

\begin{document}
\maketitle
\begin{abstract}
Empathetic dialogue is a human-like behavior that requires the perception of both affective factors (e.g., emotion status) and cognitive factors (e.g., cause of the emotion). Besides concerning emotion status in early work, the latest approaches study emotion causes in empathetic dialogue. These approaches focus on understanding and duplicating emotion causes in the context to show empathy for the speaker. However, instead of only repeating the contextual causes, the real empathic response often demonstrate a logical and emotion-centered transition from the causes in the context to those in the responses. In this work, we propose an emotion cause transition graph to explicitly model the natural transition of emotion causes between two adjacent turns in empathetic dialogue. With this graph, the concept words of the emotion causes in the next turn can be predicted and used by a specifically designed concept-aware decoder to generate the empathic response. Automatic and human experimental results on the benchmark dataset demonstrate that our method produces more empathetic, coherent, informative, and specific responses than existing models.
\end{abstract}
\section{Introduction}
\label{sec:intro}
Empathetic dialogue aims to understand the human emotional status and generate appropriate responses. Previous works have demonstrated that empathetic dialogue systems can effectively improve user experience and satisfaction in various domains, such as chit-chat \cite{26}, customer service \cite{role, lin2022duplex, he2022galaxy}, and psychological counseling \cite{27}. In psychology, two primary forms of empathy are affective empathy and cognitive empathy, constituting the ideal empathy \cite{45}. Affective empathy seeks to feel the same emotions as others, and cognitive empathy seeks to stand in someone else's situation and better understand their contextual experiences related to emotions. In empathetic dialogue research, affective empathy has been well studied, including mixture of experts \cite{18}, emotion mimicry \cite{19}, and multi-resolution user feedback \cite{20}. Cognitive empathy has gradually attracted the attention of scholars in recent years, including the emotion cause of the context \cite{41, 47}, external knowledge \cite{16, 21}, etc. 

\begin{figure*}[htp!]
  \centering
  \includegraphics[width=\textwidth]{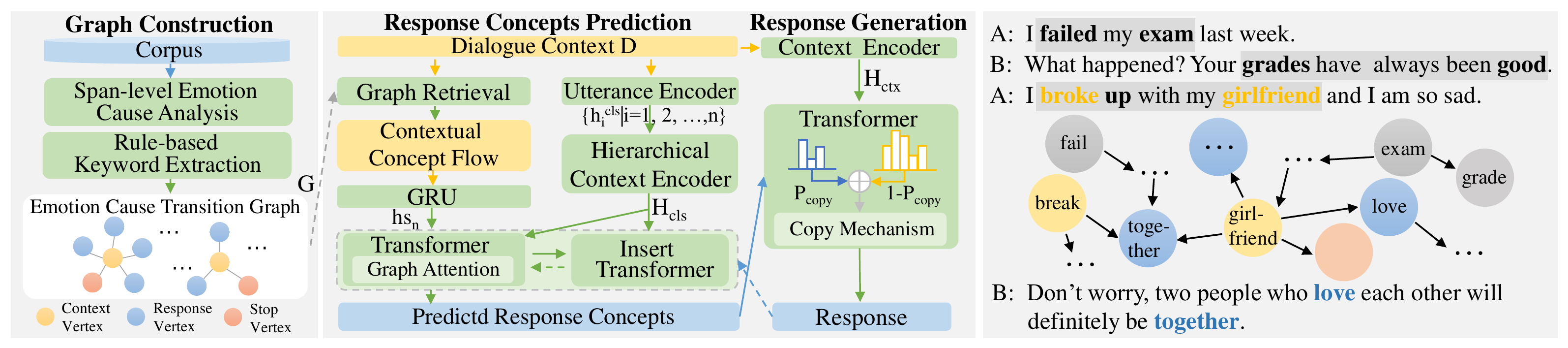}
  \caption{The overall architecture of our proposed ECTG. The left side is the model part, and the right side is the support example.}
  \label{fig:2}
\vspace{-2mm} 
\end{figure*}

As an important cognitive factor, the causes of the emotion status is an integral part of human sentiment analysis \cite{emotions, hu2022unimse}. 
However, the existing empathetic dialogue methods concerning emotion causes mainly focus on causes in the current dialogue context \cite{41, 47}. These approaches aim to understand and duplicate emotion causes in the context to show empathy for the speaker. In fact, instead of only repeating contextual causes, the real empathetic responses often demonstrate a logical and emotion-centered transition from causes in the context to those in the responses. One way to augment the emotion cause transition modeling for response generation is to introduce external knowledge with commonsense knowledge graph \cite{16, 21}. However, the transitions of emotion causes in empathetic dialogue are often emotion-centered, which are relatively sparse or absent in the commonsense knowledge graph and difficult to be effectively searched. An example is shown on the right of Figure~\ref{fig:2}. The transition from ``girlfriend'' to ``love'' and ``together'' is beyond the causes in the context and is difficult to be predicted only by the current context.

To address these issues, we propose a method, named \textbf{ECTG}, to guide the generation of empathetic responses with a \textbf{E}motion \textbf{C}ause \textbf{T}ransition \textbf{G}raph. As shown in Figure~\ref{fig:2}, the proposed method consists of three stages: Graph Construction, Response Concepts Prediction, and Response Generation. The emotion cause transition graph is automatically constructed on the golden empathetic dialogue corpus, which consumes much cost and is essential in improving empathetic dialogue. We first manually annotate a span-level emotion cause dataset and exploit a pre-trained model fine-tuned on this dataset to identify emotion cause spans. Since human dialogue \cite{zhang-etal-2022-slot, dai2022cgodial, dai2021preview} naturally centers on key concepts \cite{31, 32}, we extract keywords in emotion cause spans as key concepts, which are vertices of the graph. And edges in the graph represent natural transitions between emotion causes in the dialog. Then, combined with the hierarchical context encoder and the contextual concept flow retrieved from the graph, we use the Transformer with graph attention and Insertion Transformer to jointly optimize to predict response concepts. Finally, with the dialogue context and predicted concepts, a transformer decoder with the copy mechanism explicitly generates final responses. 
Our contributions are summarized as follows: 1) We propose a novel approach to empathetic dialogue in line with the psychology theory and the human dialogue pattern, which can effectively improve the empathetic response generation. 2) Automatic and human evaluations show that our method generates more empathetic, coherent, informative, and specific responses than existing models. 3) To extract emotion causes more accurately, we crowdsource annotated a span-level emotion cause dataset. We will publicly release the dataset for future research.

\section{METHODOLOGY}
\label{sec:method}
Formally, the constructed emotion cause transition graph is defined as $G$, given the dialog context $D$ with $n$ utterances, i.e., $D= \{U_1, U_2, … , U_n \}$, $U_i$ represents the i-th utterance in $D$. Ultimately, we aim to produce empathetic, coherent, informative, and specific responses $R$.

\subsection{Graph Construction}
To construct the emotion cause transition graph,  we first conduct the span-level emotion cause analysis. The emotion cause span is the consecutive sub-sequence of an utterance that expresses the cause of the emotion \cite{12}. 
Due to the absence of public span-level emotion cause annotated dataset for empathetic dialogue, we follow the same setting in \cite{41} and manually annotate the emotion cause spans in the dataset (Section~\ref{sec:3.1}). 

To identify emotion cause spans, we exploit pre-trained span-level SpanBERT \cite{34} to encode the dialog context and corresponding emotion label. We concatenate embeddings of the dialog context and emotion with the special token \texttt{[SEP]} as the input for the encoder. Then, we adopt Pointer network \cite{33} to generate start and end positions of spans following \cite{12}. We utilise the attention mechanism for each emotion cause span to measure the probability of different positions.

We identify the emotion cause span of each utterance in the dialog with the previous method. 
Then, we use a rule-based keyword extraction method \cite{35} to obtain significant keywords from emotion cause spans. All the extracted keywords are regarded as emotion cause concepts, which are defined as the vertices of the graph $G$. We connect two concepts with a direct edge if one concept appears in the last utterance of the context, which is the head vertex of the edge, and the other concept appears in the response, which is the tail vertex of the edge. We use point-wise mutual information (PMI)    between the head and tail vertex to filter out low-frequency concept pairs.


\subsection{Response Concepts Prediction}
To generate empathetic responses, we predict response concepts using the emotion cause transition graph. Given the i-th utterance $U_i$, all the concepts in $U_i$ which are also involved in the graph $G$ form a concept set $cs_{i}=\{c_{1}, c_{2}, \cdots, c_{m_{i}}\}$, where $m_{i}$ is the number of concepts in $U_i$. 

\textbf{Context Encoding.}  
To better utilize the dialog context \cite{he2022space2, he2022space3} in predicting response concepts, we encode the context hierarchically to collect all utterance representations. We prepend a special token \texttt{[CLS]} of each utterance $U_i$, and transform them into a sequence of hidden vectors with a BERT encoder: $h_{i}^{cls} = \text{BERT}_{\text{enc}}(\text{[CLS]}, U_i)[0]$. 

$h_{i}^{cls}$ is the hidden representation of \texttt{[CLS]}, which denotes the global memory of the utterance $U_i$. We input all $h_{i}^{cls}$ into a Transformer encoder to model the global semantic dependency between utterances:$H_{cls} =\operatorname{Trs}_{enc}\left(\left[h_{1}^{cls},
h_{2}^{cls},\cdots,h_{n}^{cls}\right]\right)$.

Then, we exploit a GRU unit to recursively encode concept sets in the dialogue context:
\begin{equation}
h s_{i} =\operatorname{G R U} \left(h s_{i-1}, \sum_{j=1}^{m_{i}} \alpha_{i j} e^{c}_{i j}\right), i \in[1, n],
\end{equation}
\begin{equation}
\alpha_{i j} =\frac{\exp \left(\beta_{i j}\right)}{\sum_{k=1}^{m_{i}}\left(\beta_{i k}\right)}
,
\beta_{i j} = hs_{i-1}^\mathsf{T} W_{3} e^{c}_{i j},
\end{equation}
where $e^{c}_{i j}$ is concept embedding, $hs_{i}$ represents contextual concept flow. $\alpha_{i j}$ is used to measure the probability of transitions to associated concepts.

\textbf{Response Concepts Selection.} We combine the dialogue context representation and the previously decoded concepts by a Transformer decoder, as a basis for dynamically selecting the next vertex in the emotion cause transition graph:
$hdc_{t}=\operatorname{Trs_{dec}}\left(\left[e^{dc}_{1:t-1}\right], H_{cls}\right)$.
Here, $e^{dc}_{1:t-1}$ denotes the embeddings of previously decoded concepts at step t.

For the concept set $cs_{n}$ of the last utterance $U_{n}$ in the context, we retrieve a group of subgraphs in the graph $G$, where each concept in $cs_{n}$ is the head vertex and its each direct neighbor vertex is the tail vertex. The subgraph $g_{i}=\left\{\left(c_{j}, c_{jk}\right)\right\}_{k=1}^{N_{j}}, c_{j} \in cs_{n}$, where $N_j$ is the number of vertex pairs of $c_{j}$ in $g_i$. We employ a dynamic graph attention mechanism to calculate the subgraph vector:
\begin{equation}
    \alpha_{j} =\frac{\exp \left(\beta_{j}\right)}{\sum_{l=1}^{m_{i}} \exp \left(\beta_{l}\right)},
\end{equation}
\begin{equation}
\beta_{j} =\left(W_{4}\left[hdc_{t} ; hs_{n}\right]\right)^\mathsf{T} \cdot \left(W_{5}\sum_{k=1}^{{N}_{j}} \alpha_{j k}\left[e_{j}^{c} ; e_{j k}^{c}\right]\right),
\end{equation}
where $\alpha_{j}$ determines the choice of subgraphs. $ hs_{n} $ incorporates information of contextual concept flow. $\alpha_{j k}$ determines which tail vertex is selected in $g_i$:
\begin{equation}
\alpha_{j k} =\frac{\exp \left(\beta_{j k}\right)}{\sum_{l=1}^{N_{j}} \exp \left(\beta_{j l}\right)},
\end{equation}
\begin{equation}
\beta_{j k}=(W_{6} [hdc_{t} ; hs_{n} ; e^{c}_{j}])^\mathsf{T} \cdot W_{7} e^{c}_{j k}).
\end{equation}

Finally, the chosen response concepts at step t are derived as:
$P(dc_{t} \mid D, G, dc_{<t})=\alpha_{j} \cdot \alpha_{j k}$.

\textbf{Response Concepts Refining.} 
From the pilot study, we found that the response concept decoder pays more attention to frequent concepts and thus lacks variety. We conjecture that supervision signals are only concept labels but the signals from the natural empathetic response should also be used simultaneously to optimize the decoder. To solve this issue, we propose an auxiliary module that takes intermediate layers of the response concept decoder as input and takes the empathetic response as output, and optimizes with the response concept decoder together via multi-task learning. In this way, the information of empathetic responses can be transported into the response concept decoder to facilitate more abundant response concept prediction. More specifically, we exploit the Insertion Transformer \cite{39} in a non-autoregressive manner as the auxiliary loss to choose predicted concepts inspired by \cite{32}. 
The loss of the Insertion Transformer is:
$
\operatorname{L_{g}}=\frac{1}{k+1} \sum_{pos=0}^{k} \sum_{n=il}^{jl}-\log P^{\text{InsTrs}}_{n} \cdot w_{pos}(n)
$. For more details about the Insertion Transformer, please refer to \cite{39}.

For the loss of response concepts $C$, we use negative log-likelihood loss:
$L_{c}=\frac{1}{|C|} \sum_{t=1}^{|C|}-\log p(c_{t} \mid D, G, c_{<t})$.

The optimization of predicted concepts that can generate the empathetic response is determined by the weighted sum of two previous losses:
$\text{Loss}_{gc}=L_{g}+r L_{c}$.
Here, $r$ is the coefficient to control the impact of concept loss.

\subsection{Empathetic Response Generation}
To generate the empathetic response, we concatenate the predicted response concepts and the previous dialog context together as a sequence to the BERT encoder, and then combine a Transformer decoder with the copy mechanism to explicitly utilize it. The final generation probabilities are computed over the word vocabulary and the selected concept words:
\begin{equation}
H_{ctx} =\operatorname{BERT}_{enc} \left(input_{D^{c}}\right), H_{dec} =\operatorname{Trs}_{dec}\left(H_{ctx}\right),
\end{equation}
\begin{equation}
P(w) = A_{h} \odot P_{copy} \cdot M_{src} + (1-P_{copy})P_{gw}(w),
\end{equation}
\begin{equation}
P_{copy} = \operatorname{Sigmoid}(W_{8}\cdot H_{dec}),
\end{equation}
\begin{equation}
P_{gw}(w) = \operatorname{Softmax}(W_{9} \cdot H_{dec}),
\end{equation}
where $D^{c}$ is the input combining the dialogue context and predicted concepts, $input_{{D}^{c}}$ is the input ids of $D^{c}$. $P_{copy}$ is the probability of copying a particular word from the attention distribution directly, $M_{src}$ is an indicator matrix mapping each source word to the additional vocab containing it. We apply the cross-entropy loss for training.

\begin{table*}[htp]
\centering

\resizebox{0.68\textwidth}{20mm}{
\begin{tabular}{lcccccc|ccccc}
\toprule
 \multirow{2}{*}{Models} & \multicolumn{6}{c|}{Automatic Evaluation} & \multicolumn{5}{c}{Human Evaluation} \\

\midrule
    Multi-trs   & 2.103   & 0.1948    & 0.456   & 1.947 & 16.67 & 12.81 & 2.91    & 2.87    & 2.48    & 4.86    & 0.24    \\
    MoEL  & 1.933   & 0.2166    & 0.469   & 2.155   & 17.00 & 14.60 & 2.89    & 2.87    & 2.46    & 4.93    & 0.21    \\
    MIME  & 1.894   & 0.2039    & 0.449   & 1.829  & 16.64 & 13.68 & 3.13    & 2.97    & 2.59    & 4.89    & 0.24    \\
    EmpDG    & 1.975   & 0.2188    & 0.470   & 1.981 & 17.34 & 14.70 & 3.00    & 2.97    & 2.55    & 4.93    & 0.24    \\
    EC (soft)     & 1.345   & 0.1925    & 1.698   & 8.493 & 15.67 & 10.21 & 2.96    & 3.00    & 2.53    & 4.92    & 0.27    \\
    KEMP  & 1.762   & 0.1948    & 0.660   & 3.074 & 15.43 & 12.78 & 2.78    & 2.72    & 2.46    & \textbf{4.94} & 0.21    \\
    CEM   & 1.629   & 0.2134    & 0.645   & 2.856 & 16.27 & 15.83 & 3.02    & 3.21    & 2.38    & 4.90    & 0.19    \\
    DialoGPT  & 0.734   & 0.1515    & \textbf{3.140}   & \textbf{17.551}   & 8.51 & 7.00 &3.70    & 3.89    & 3.06    & 4.88    & 0.61 \\
\midrule
   ECTG  & \textbf{5.467}   & \textbf{0.2701} & 1.840   & 16.404 & \textbf{23.77} & \textbf{51.43} & \textbf{3.78}\textsuperscript{$\ddagger$} & \textbf{4.13}\textsuperscript{$\ddagger$} & \textbf{3.13}\textsuperscript{$\ddagger$} & 4.88    & \textbf{0.64}\textsuperscript{$\ddagger$} \\
   
\bottomrule
\end{tabular}
}
\vspace{-1mm}
\caption{Automatic and human evaluations. $\dagger$, $\ddagger$ represent the statistical significance (t-test) with p-value \textless 0.05 and 0.01.}
\label{tab:mexp}
\vspace{-4mm} 
\end{table*}

\section{EXPERIMENTS}
\label{sec:experiments}

\subsection{Experimental Setup}
\label{sec:3.1}
\textbf{Datasets \& Evaluation Metrics.} We conduct experiments on the EmpatheticDialogues \cite{22},
which is a large-scale English multi-turn empathetic dialogue benchmark dataset. 
For automatic metrics, we adopt BLEU-4 (B-4), BERTscore F1 ($\text{F}_{\text{BERT}}$) \cite{50}, Distinct-n (Dist-1/2), ROUGE-L (R-L), CIDEr to evaluate the performance of response generation. For human evaluation, we randomly sample 100 dialogues from testing set and employ crowdsourcing workers to rate generated responses based on five aspects of Empathy, Coherence, Informativity, Fluency, and Specificity. The score is from 1 to 5 (1: not at all, 3: OK, 5: very good), except Specificity. The Specificity score is 1 or 0, representing yes or no. Fleiss' Kappa of the human evaluation results is 0.498, indicating moderate agreement. 

\textbf{Baselines \& Hyper-parameters.} We choose MoEL \cite{18}, MIME \cite{19}, EmpDG \cite{20}, EC (soft) \cite{47}, KEMP \cite{21}, CEM \cite{16}, and DialoGPT (345M) \cite{51} as baselines. For vertices in the graph, we use VGAE \cite{40} to initialize representations, and the embedding size is 128. The hidden size of GRU is 768, and the maximum number of concepts is 5. We use Adam for optimization with the initial learning rate of 0.001.


\subsection{Results and Analysis}
\textbf{Automatic and Human Evaluations.}
Table~\ref{tab:mexp} reports the automatic and human experimental results. We observe that ECTG considerably exceeds baselines in most metrics for the automatic evaluation, demonstrating that ECTG is beneficial for empathetic dialogue generation. 
ECTG also achieves the best performance in four aspects for the human evaluation except Fluency, which verifies that ECTG can generate more empathetic, coherent, informative, and specific responses with the guidance of emotion causes and the transition of concepts. Additionally, we note that there is no significant difference in Fluency between models, and we speculate that the responses generated by all models are already fluent.

\begin{table}[htb]
  \centering
  \resizebox{0.8\columnwidth}{!}{
  \begin{tabular}{lcccccc}
    \toprule
    \multirow{1}{*}{Models} & B-4 & $\text{F}_{\text{BERT}}$ & Dist-1 & Dist-2 & R-L & CIDEr\\
    \midrule
    ECTG & \textbf{5.47} & \textbf{0.2701} & 1.84 & \textbf{16.40} & \textbf{23.77} & \textbf{51.43}\\
    \hline
    w/o copy & 2.75 & 0.2569 & \textbf{2.37} & 14.73 & 20.95 & 39.62\\
    w/o seca & 3.04 & 0.2539 & 2.27 & 14.22 & 21.32 & 40.73  \\
    w/o graph & 3.21 & 0.2301 & 2.18 & 13.56 & 18.88 & 23.98 \\
  \bottomrule
\end{tabular}
}
\vspace{-1mm}
\caption{Results of the ablation study.}
\label{tab:ablation}
\vspace{-5mm}
\end{table}

\textbf{Ablation Study.}
We designed three variants of ECTG for the ablation study: \textbf{1) w/o copy}. We remove the Transformer decoder with the copy mechanism and only employ the non-autoregressive generation. \textbf{2) w/o seca}. The span-level emotion cause analysis is removed, then all keywords in the utterance are adopted to construct the graph. \textbf{3) w/o graph}. We remove the emotion cause transition graph and replace it with the form of text. The obtained results are shown in Table~\ref{tab:ablation}. We can observe that variants drop dramatically in most metrics, indicating our model settings' effectiveness. According to statistics, responses generated by ECTG tend to be longer than those generated by variants. It may have a great impact when calculating uni-gram. However, other metrics help prove that responses generated by ECTG are better.

\textbf{Case Study.}
In Table~\ref{tab:case1}, we provide some cases to compare generated responses of ECTG and baselines. 
In the first case, affective empathy oriented baselines roughly perceive the user's emotion status and respond generally. Although models with additional knowledge convey more information, their responses are not targeted to the context. EC(soft) successfully identifies the user's emotional state and replies with specific examples. However, the response is not particularly coherent due to the lack of global graph guidance. In contrast, ECTG understands the user's emotions and experiences accurately and gives good wishes with empathetic, relevant, and non-universal responses. In the second case of multiple-turn dialogue context, compared with other models that acknowledge the user's emotion, ECTG expresses appropriate emotion and explores more valuable information.

\begin{table}[htb]
  \resizebox{\columnwidth}{!}{
  \begin{tabular}{l|l}
    \toprule
  Emotion & Nostalgic  \\
   Context & \makecell[tl]{"I recently \sethlcolor{lightgray}\hl{\textbf{spoke} with my \textbf{ex-girlfriend} on the phone. The \textbf{conversation}}\\ \sethlcolor{lightgray}\hl{went pretty \textbf{well}, and it \textbf{reminded} me of my \textbf{past experiences} with her}."} \\ 
   \midrule
   Multi-trs & \makecell[tl]{That is so sweet. I am glad you were able to see her.} \\
   MoEL & \makecell[tl]{That is so nice of you. I am glad you are okay.} \\
   MIME & \makecell[tl]{That is nice of you. I am sure you miss her a lot of hard work.} \\
   EmpDG & \makecell[tl]{That is very nice of you.} \\
   EC (soft) & \makecell[tl]{Oh, that's so sweet. I'm sure you were very grateful for her.} \\
   KEMP & \makecell[tl]{That is great. I am sure you will have a great time to have a blast.} \\
   CEM & \makecell[tl]{That is great. Did you find out of it?} \\
   DialoGPT & \makecell[tl]{I hope everything turned out well for you.} \\
   \midrule
   ECTG & \makecell[tl]{That's \textcolor{red}{interesting}. I \textcolor{red}{hope} it turns out well for you two can still be \textcolor{red}{together}.} \\
    \midrule
    \midrule
  Emotion & Surprised \\
   Context & \makecell[cl]{"I couldnt celebrate my 18th birthday as I had exams during the time \\and I felt terrible for it. \\But \sethlcolor{lightgray}\hl{my \textbf{friends} threw a \textbf{surprise party} for me} and it was the best!"\\
   "That's \textbf{great}! How did they \textbf{surprise} you?"\\
   "I was studying in my bedroom for a couple of hours straight with headphones\\ on and then my mom called me to the living room for dinner and when \sethlcolor{lightgray}\hl{I went}\\ \sethlcolor{lightgray}\hl{out the \textbf{whole place} was \textbf{decorated} and \textbf{everyone was there}}."}\\
   \midrule
  Multi-trs & \makecell[tl]{That is good. I am glad you are able to do that.} \\
   MoEL & \makecell[tl]{That is so nice of them!} \\
   MIME & \makecell[tl]{That is a good thing to do. I am sure you are going to have a good time.} \\
   EmpDG & \makecell[tl]{That is good. I am sure you will be able to get a little one.} \\
   EC (soft) & \makecell[tl]{That's awesome! I hope you get the next step!} \\
   KEMP & \makecell[tl]{I am sure you will do great!} \\
   CEM & \makecell[tl]{That is great! I love the day!} \\
   DialoGPT & \makecell[tl]{That's so nice of them!} \\
   \midrule
   ECTG & \makecell[tl]{That's so \textcolor{red}{sweet} of them! Did you \textcolor{red}{enjoy} it?} \\
  \bottomrule
\end{tabular}
}
\caption{Cases of responses generated by all models. The identified spans are marked with shadow. The concepts in the context are bold, and the predicted concepts are colored red.}
\vspace{-1mm}
\label{tab:case1}
\vspace{-2mm}
\end{table}

\textbf{Exploration Experiment.}
We further explore the transferability of ECTG concepts by integrating predicted response concepts into the pre-trained model as prompts. We adopt a large-scale dialogue model DialoGPT (345M), whose parameter number is significantly higher than our model. We also choose BlenderBot \cite{25} as the reference for the pre-trained model in the field of empathetic dialogue, which is trained with multiple communication skills. The results in Table~\ref{tab:explore} show that DialoGPT with concepts of ECTG outperforms DialoGPT and BlenderBot in most metrics, which verifies that combining predicted response concepts can improve performance.

\begin{table}[htp]
  \centering
  \resizebox{0.85\columnwidth}{!}{
  \begin{tabular}{lcccccc}
    \toprule
    \multirow{1}{*}{Models} & B-4 & $\text{F}_{\text{BERT}}$ & Dist-1 & Dist-2 & R-L & CIDEr\\
    \midrule
    BlenderBot & 1.30 & \textbf{0.1599} & 2.41 & 12.45 & 9.71 & 8.07 \\
    \midrule
    DialoGPT & 0.73 & 0.1515 & 3.14 & 17.55 & 8.51 & 7.00\\
    + ECTG concepts & \textbf{1.36} & 0.1524 & \textbf{3.75} & \textbf{21.39} & \textbf{10.78} & \textbf{15.91}\\
  \bottomrule
\end{tabular}
}
\vspace{-1mm}
\caption{Results of the exploration experiment.}
\label{tab:explore}
\vspace{-4mm} 
\end{table}


\section{CONCLUSION}
\vspace{-1mm}
\label{sec:con}
In this paper, we propose to generate empathetic responses aware of emotion cause concepts. We construct an emotion cause transition graph to explicitly model natural transitions in the human empathetic dialogue and design a model using the graph to benefit the empathetic response generation. Automatic and human evaluations verify our approach's ability in the field of empathetic dialogue. 

\vfill\pagebreak




\bibliographystyle{IEEEbib}
\small
\bibliography{refs}

\end{document}